\documentclass[journal,10pt]{IEEEtran}

\usepackage{amsmath,amssymb}
\usepackage{booktabs,multirow,array,tabularx}
\usepackage{graphicx}
\usepackage[table]{xcolor}
\usepackage{tikz}
\usetikzlibrary{arrows.meta,positioning,fit,calc,backgrounds}
\usepackage{pgfplots}
\usepgfplotslibrary{groupplots}
\pgfplotsset{compat=1.18}
\pgfplotsset{
  seedone/.style={paceorange,solid,mark=*},
  seedeighteen/.style={paceblue,dashed,mark=square*},
  seedtsix/.style={pacegreen,dashdotted,mark=triangle*},
  seedftwo/.style={black!65,densely dotted,mark=diamond*},
  seedotthree/.style={red!70,solid,mark=o},
  seedfsix/.style={purple!70,dashed,mark=square},
  seednnine/.style={brown!70,dashdotted,mark=triangle},
  seedttsix/.style={cyan!70!black,densely dotted,mark=diamond}
}
\usepackage{cite}
\usepackage{microtype}
\usepackage{placeins}
\usepackage{float}
\usepackage{needspace}
\usepackage{flushend}
\usepackage{hyperref}

\hypersetup{
  hidelinks,
  pdftitle={PACE-MoE: Adaptive Domain Sampling and Domain-Specific LoRA for Medical Visual Grounding},
  pdfauthor={Zhenxuan Wang}
}
\renewcommand{\arraystretch}{1.10}
\newcommand{\method}{PACE-MoE}

\newcommand{\dataset}{U-MRG-6K}
\definecolor{paceblue}{RGB}{39,88,148}
\definecolor{pacegreen}{RGB}{53,126,93}
\definecolor{paceorange}{RGB}{184,103,31}
\definecolor{tableblue}{RGB}{234,242,247}
\definecolor{tablebluehead}{RGB}{218,233,244}
\definecolor{tableyellow}{RGB}{255,244,204}
\definecolor{tableyellowstrong}{RGB}{255,235,166}
\definecolor{tablegreen}{RGB}{232,247,237}
\definecolor{tablegray}{RGB}{246,246,246}
\definecolor{checkgreen}{RGB}{24,140,73}
\definecolor{crossred}{RGB}{196,49,54}
\newcommand{\cmark}{\textcolor{checkgreen}{\ensuremath{\checkmark}}}
\newcommand{\xmark}{\textcolor{crossred}{\ensuremath{\times}}}

\providecommand{\IEEEmembership}[1]{\textit{#1}}

\begin{document}

\title{PACE-MoE: Adaptive Domain Sampling and Domain-Specific LoRA for Medical Visual Grounding}
\date{}

\author{Zhenxuan~Wang, Jisheng~Dang,  Ronghao~Lin, Bimei~Wang, Jingwen Zhao, Wenjun Ma,
Bin~Hu,~\IEEEmembership{Fellow, IEEE},\\
Qi~Tian,~\IEEEmembership{Fellow, IEEE}, and Tat-Seng~Chua%
\thanks{This work was supported by the National Natural Science Foundation of
China (Grants No. 62227807 and U24B20186) and the Supercomputing Center of
Lanzhou University.}%
\thanks{Zhenxuan Wang, Jisheng Dang,  Bimei Wang, Jingwen Zhao, Wenjun Ma are with the
School of Information Science \& Engineering, Lanzhou University, Lanzhou
730000, China (e-mail: dangjsh@mail2.sysu.edu.cn).}%
\thanks{Ronghao Lin is with the College of Computer Science and Software
Engineering, Shenzhen University, Shenzhen 518060, China (e-mail:
linrh@szu.edu.cn).}%
% \thanks{Wei-Shi Zheng is with the School of Computer Science and Engineering,
% Sun Yat-sen University, Guangzhou 510006, China (e-mail: wszheng@ieee.org).}%
\thanks{Bin Hu is with the School of Medical Technology, Beijing Institute of
Technology, Beijing 100081, China (e-mail: bh@bit.edu.cn).}%
\thanks{Qi Tian is with Cloud and AI BU, Huawei, Shenzhen, Guangdong 518129,
China (e-mail: tian.qi1@huawei.com).}%
\thanks{Tat-Seng Chua is with the School of Computing, National University of
Singapore, Singapore 119077 (e-mail: dcscts@nus.edu.sg).}%

\thanks{Corresponding authors: Jisheng Dang, Ronghao Lin,  and Bin Hu.}}

\markboth{IEEE Transactions on Multimedia, Under Review}%
{Wang: Adaptive Domain Sampling for Medical Visual Grounding}

\maketitle

\begin{abstract}
Medical visual grounding requires models to align clinical descriptions with spatial evidence across heterogeneous imaging domains. However, existing multi-domain post-training strategies typically rely on fixed domain sampling and shared parameter updates, overlooking the need to adapt both which domains receive training and where their updates are applied. To address this issue, we propose \textbf{PACE-MoE}, a policy-guided framework that jointly performs adaptive domain sampling and domain-specific adaptation for medical visual grounding. PACE dynamically assesses the training value of each imaging domain from current-policy feedback and progressively adjusts domain coverage through domain admission, replay, and final all-domain balancing. Meanwhile, domain-specific LoRA adapters route updates to the corresponding imaging domains while preserving the pretrained backbone and visual encoder. This coordinated design aligns training-data allocation with parameter adaptation without modifying the original reward formulation or Group Relative Policy Optimization (GRPO) objective. Experiments across seven medical imaging domains show that PACE-MoE consistently improves grounding performance over fixed-domain training, achieving a 3.64\% relative improvement in mean box IoU over the Vanilla baseline. Under matched domain-specific adapter architectures, adaptive sampling alone further improves mean box IoU by 2.07\%, demonstrating the benefit of policy-guided domain allocation. These results establish coordinated data–parameter routing as an effective strategy for multi-domain medical visual grounding.
\end{abstract}

\begin{IEEEkeywords}
Medical visual grounding, multimodal large language model, mixture of experts,
low-rank adaptation, reinforcement learning, curriculum learning.
\end{IEEEkeywords}

\Needspace{5\baselineskip}
\section{Introduction}
\IEEEPARstart{M}{edical} visual grounding requires a model to connect a
clinical description with the corresponding image region. A query may refer
to a pathology, an anatomical relation, or a diagnostic clue without naming
the target directly. The model must therefore combine language understanding
with spatial localization. This requirement extends across CT, MRI,
radiography, ultrasound, endoscopy, dermoscopy, and mammography, despite their
different appearances and acquisition processes.

MedReasoner~\cite{yan2026medreasoner} addresses this problem with a clinical
reasoning model that emits a reasoning trace, a box, and two points; frozen
MedSAM2~\cite{ma2025medsam2} converts the prompts into a mask. Because format
and spatial overlap are directly verifiable, the reasoner can be optimized
with Group Relative Policy Optimization (GRPO)
\cite{shao2024deepseekmath,guo2025deepseekr1}. However, improvements from
post-training also depend on how heterogeneous training data are used. We
treat each imaging modality as a domain with its own data
distribution~\cite{zhou2023dgsurvey}. All seven domains and their labels are
available during both training and evaluation. The setting is thus
multi-domain learning, rather than generalization to unseen domains.

A useful training mixture should reflect the current
model, not only the number of examples in each domain. In GRPO, responses to
the same query are compared through their rewards. Malformed responses or
nearly identical rewards can provide little useful spatial feedback even
when many examples are available. As the model changes, the domains that
supply usable feedback can also change. Fixed sampling does not capture this
dependence on the learning state.

\begin{figure*}[t]
\centering
% Keep PACE_MoE_Fig1_motivation.pdf in the Overleaf project root.
\includegraphics[width=\textwidth]{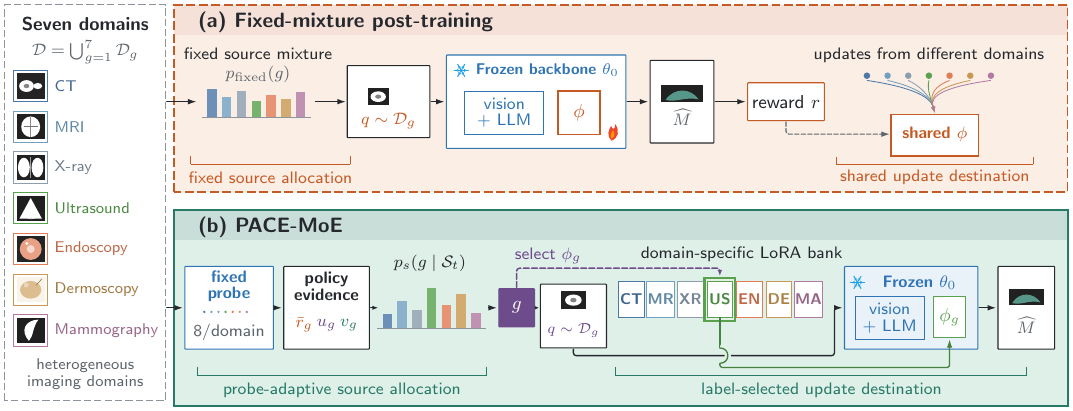}
\caption{Motivation for domain-aware post-training. (a) A fixed data mixture
does not respond to the current model, and one shared LoRA adapter receives
updates from all domains. (b) \method{} uses policy feedback to adapt the
sampling distribution $p_s(g\mid\mathcal S_t)$. The selected domain label
$g$ also identifies the adapter to update. The pretrained backbone remains
shared and frozen.}
\label{fig:motivation}
\end{figure*}

Data allocation is only one part of the problem. A shared LoRA adapter
receives updates from every imaging domain, although the domains may require
different adjustments to the pretrained model. Domain-specific adapters
retain the common backbone while giving each domain its own trainable
parameters. Yet this separation does not determine how often an adapter
should be trained. Conversely, adaptive sampling alone does not change which
parameters are shared. Figure~\ref{fig:motivation} motivates considering
these two decisions together: allocating queries across domains and assigning
their updates to the corresponding adapters.

We propose \method{} to coordinate domain sampling
and domain-specific adaptation using the current policy as feedback. A fixed
probe set estimates whether each domain provides valid responses and useful
reward contrast. The resulting curriculum adjusts domain coverage over
training instead of committing to an initial ordering. Replay retains earlier
domains, and a final all-domain stage gives additional attention to domains
with lower probe reward.

The sampled domain label also selects the LoRA adapter used to generate and
learn from its responses. Thus, the curriculum determines the training
frequency of each adapter, and the adapted model supplies the next round of
feedback. Building on established domain-specific
adaptation~\cite{omi2022multidomain}, this design uses the known labels without
learning another routing network. The dense backbone, visual encoder, and
MedSAM2 remain frozen; the reward and GRPO objective are unchanged. The
framework adapts how training data and domain-specific parameters are used
while retaining the released grounding pipeline.

The contributions are summarized as follows:
\begin{itemize}
  \item We introduce a domain-aware post-training framework that links
  training-data allocation to domain-specific model updates. It retains
  shared pretrained representations while assigning each domain's responses
  to its corresponding adapter.
  \item We develop a policy-guided curriculum that adapts domain coverage
  to the current model. It combines feedback-based admission with replay and
  final balancing, so that prioritizing useful feedback does not remove
  earlier domains from subsequent training.
  \item We evaluate the complete framework and its sampling component
  separately. The full system improves mean box IoU over Vanilla by 3.64\%
  in relative terms. Within-architecture controls assess adaptive sampling
  across eight shared-adapter seeds and five domain-specific-adapter seeds;
  ablations examine the roles of replay and final balancing.
\end{itemize}

\section{Related Work}

\subsection{Multi-Domain Visual Learning}

A domain is a data distribution associated with an acquisition setting,
environment, or population. Unlike domain generalization to unseen target
domains~\cite{zhou2023dgsurvey}, our setting contains the same seven imaging
domains during training and evaluation. Multi-domain methods often combine
shared features with domain-specific components~\cite{omi2022multidomain}.
We study this design together with policy-guided domain sampling for GRPO
grounding.

\subsection{Reasoning-Based Medical Visual Grounding}

Medical MLLMs extend vision-language modeling to clinical terminology and
heterogeneous imaging. MPMA jointly models medical image--text reconstruction
and cross-modal alignment to learn representations with domain knowledge
\cite{zhang2024mpma}. LLaVA-Med uses biomedical visual instruction tuning
\cite{li2023llavamed}, while Lingshu targets unified medical understanding and
reasoning \cite{xu2025lingshu}. Language-guided segmentation methods include
LISA \cite{lai2024lisa}, Seg-Zero \cite{liu2025segzero}, and MedGround-R1
\cite{xu2025medgroundr1}. MedReasoner separates reasoning from mask decoding:
a trainable reasoner predicts spatial prompts and a frozen segmentation expert
produces the mask \cite{yan2026medreasoner}. We retain this output and reward
pipeline while changing domain sampling and the assignment of LoRA adapters.

\subsection{Spatial Representations and Grounded Visual Reasoning}

General visual segmentation offers complementary approaches to maintaining
spatial evidence. Isogenous memory sampling with frame-relation mining
addresses memory matching and temporal context~\cite{dang2023imsfr}.
Multi-frame context memory captures information across adjacent
frames~\cite{dang2024beyond}, while adaptive sparse memory selects
informative past frames and retrieves relevant local
features~\cite{dang2025asm}. Video Decoupling Networks separate scene,
motion, and instance cues~\cite{dang2026vdn}, and Fast Track Anything uses
sparse spatio-temporal propagation for unified video
segmentation~\cite{dang2026fasttrack}. These studies examine visual
representations and temporal propagation; our focus is the allocation of
domain queries and adapter updates during medical post-training.

Visual-language reasoning also depends on selecting evidence and preserving
semantic consistency. REVU retrieves external object-action memory to
support long-term video understanding~\cite{dang2025externalmemory}.
Hierarchical multimodal consistency uses multi-level alignment to reduce
video-language hallucinations~\cite{dang2025hallucination}. Fact generates
faithful, concise, and transferable rationales through verifiable visual
programming~\cite{gao2024fact}. Diff-LMM~\cite{dang2025difflmm} and
PhysMLLMs~\cite{yan2026physmllms} further investigate teacher-guided spatial
representation learning for general visual understanding.

\subsection{Reinforcement Learning for Medical Grounding}

GRPO estimates a query-specific baseline from multiple responses, avoiding a
separate critic \cite{shao2024deepseekmath,guo2025deepseekr1}. It normalizes
rewards within a group but does not determine domain sampling or parameter
sharing. QoQ-Med addresses heterogeneity through Domain-aware Relative Policy
Optimization (DRPO), which rescales advantages using domain and cluster
statistics \cite{dai2025qoqmed}. PACE instead changes which domains supply
queries before responses are generated. Sampling and post-generation
reweighting therefore address different decisions.

\subsection{Adaptive Data Selection and Curriculum Learning}

Curriculum learning organizes examples rather than sampling from an unchanged
distribution \cite{bengio2009curriculum}; self-paced learning adapts that
organization to the learner \cite{kumar2010selfpaced}. Recent work has brought
adaptive data selection into RL-based LLM post-training. DUMP combines policy
advantages with an upper-confidence-bound sampler across data distributions
\cite{wang2025dump}, whereas SEC treats problem categories as a non-stationary
bandit and uses absolute advantage as the learning signal \cite{chen2025sec}.
Prioritized Replay updates problem priorities from empirical success rates
\cite{fatemi2026prioritized}, while Actor-Curator learns a neural curator that
selects problems to maximize policy improvement \cite{gu2026actorcurator}.
PACE scores known imaging domains on a fixed, gradient-free probe using
spatial reward, within-query dispersion, and output validity. It connects
staged domain coverage with domain-specific LoRA selection for medical
grounding.

Medical curricula use task structure, anatomical similarity, or reasoning
difficulty
\cite{rui2025medcco,song2025anatomyr1,kim2025medclm}. MedLoc-R1 adapts reward
strictness from online performance \cite{yang2026medlocr1}, while CX-Mind uses
a predefined closed-to-open curriculum for chest X-ray GRPO
\cite{li2026cxmind}. PACE leaves reward strictness unchanged and repeatedly
updates the sampling order across seven imaging domains.

\subsection{Mixture of Experts and Low-Rank Adaptation}

Med-MoE learns a router over domain-specific experts for supervised medical
VQA and classification \cite{jiang2024medmoe}. We instead use known modality
labels to select low-rank FFN adapters during GRPO grounding. MoE-LLaVA extends
sparse expert routing to large vision-language models while keeping only a
subset of experts active for each input \cite{lin2026moellava}. More generally,
mixture-of-experts models route inputs through specialized parameter subsets
\cite{shazeer2017moe}; sparse routing has also been used to scale visual
representations \cite{riquelme2021vmoe}. Replicating full experts is expensive
for a 7B multimodal model. LoRA represents an update with two low-rank matrices
while freezing the pretrained weight \cite{hu2022lora}. MoCLE routes
vision-language instructions to cluster-conditional LoRA experts
\cite{gou2023mocle}. Our Domain-MoE differs in using an observed medical
imaging-domain label, rather than an instruction cluster or a learned router,
as a hard Top-1 selection during GRPO. It provides domain-specific adaptation
without copying the complete reasoner or adding a load-balancing objective.

\section{Problem Formulation}

\subsection{Multi-Domain Medical Grounding}

Let the multi-domain training set be
\begin{equation}
  \mathcal D=\bigcup_{g=1}^{G}\mathcal D_g,
\label{eq:domain_union}
\end{equation}
where $g$ is the known imaging-domain label. Each example
$x=(I,Q,Y,g)$ contains an image $I$, a clinical query $Q$, and spatial
supervision $Y=(B^\star,P_1^\star,P_2^\star,M^\star)$. Following MedReasoner,
the policy produces
\begin{equation}
  o=(T,\hat B,\hat P_1,\hat P_2)\sim
  \pi_{\theta_0,\phi_g}(\cdot\mid I,Q),
\end{equation}
where $\theta_0$ is the frozen shared backbone and $\phi_g$ is the LoRA
adapter for domain $g$. A frozen segmenter returns
\begin{equation}
  \hat M=F_{\mathrm{seg}}(I,\hat B,\hat P_1,\hat P_2).
\end{equation}

\subsection{Group-Relative Policy Optimization}

All experiments use the released MedReasoner Soft reward, which combines
format checks with smoothed box-IoU and point-overlap terms and penalizes
invalid outputs \cite{yan2026medreasoner}. Our method introduces
neither a domain-specific reward nor an additional reward coefficient. PACE
uses the same scalar reward for domain scoring and GRPO optimization.

For a query $q=(I,Q)$, GRPO samples $K$ responses. Let $r_{q,k}$ be the
verifiable reward of response $k$. Its group-relative advantage is
\begin{equation}
  \hat A_{q,k}=\frac{r_{q,k}-\mu_q}{\sigma_q+\epsilon},
  \qquad
  \mu_q=\frac{1}{K}\sum_{j=1}^{K}r_{q,j},
\label{eq:advantage}
\end{equation}
where $\sigma_q$ is the within-group reward standard deviation. The clipped objective can be written as
\begin{align}
\mathcal J(\phi;\theta_0)=\mathbb E_{q,k,t}\big[&
  \min\{\rho_{q,k,t}\hat A_{q,k},\nonumber\\[-2pt]
  &\operatorname{clip}(\rho_{q,k,t},1-\varepsilon,1+\varepsilon)
  \hat A_{q,k}\}\nonumber\\[-2pt]
  &-\beta D_{\mathrm{KL}}(\pi_{\theta_0,\phi}\Vert\pi_{\mathrm{ref}})\big].
\label{eq:grpo}
\end{align}
Here, $\rho_{q,k,t}$ is the token-level probability ratio between the current
and rollout policies, $\varepsilon$ is the clipping threshold, and $\beta$
weights regularization toward the fixed reference policy $\pi_{\mathrm{ref}}$.

\subsection{Domain-Aware Update Allocation}

Equations~\eqref{eq:advantage}--\eqref{eq:grpo} describe what happens after a
query is selected. Standard continuation draws from fixed $p(g)$ and writes
all low-rank updates to one shared adapter. \method{} instead uses a
stage-dependent $p_s(g)$ and selects the matching adapter $\phi_g$.
Response format, reward, advantage calculation, and clipped loss remain
unchanged.
The scheduler is motivated by Eq.~\eqref{eq:advantage}. A domain with low
output validity does not yet provide reliable spatial feedback. A domain with
nearly identical response rewards provides weak within-group contrast. PACE
uses a small probe set to estimate the current training value of each domain.
The resulting score is policy-dependent; it is not a permanent difficulty
label.

\section{PACE-MoE}

\begin{figure*}[t]
\centering
% Keep PACE_MoE_Fig1_polished_v18.pdf in the Overleaf project root.
\includegraphics[width=\textwidth]{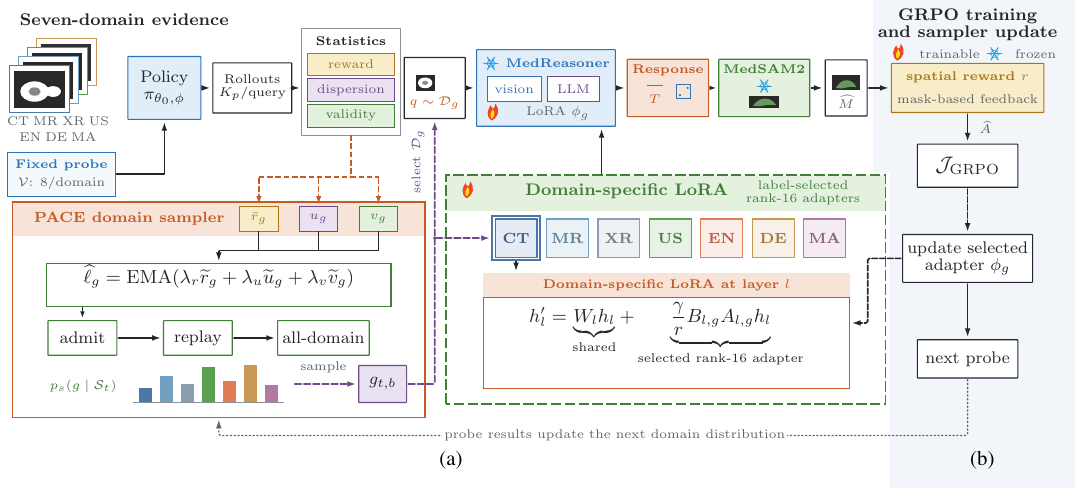}
\caption{Overview of \method{}. (a) Architecture of \method{}. Current-policy
rollouts on a fixed seven-domain probe buffer estimate reward, within-query
dispersion, and output validity. These statistics update the domain sampling
distribution through domain admission, replay, and a final all-domain stage.
The sampled label $g_{t,b}$ selects a query from $\mathcal D_g$ and activates
the corresponding rank-16 FFN LoRA adapter $\phi_g$. (b) Optimization pipeline.
The selected adapter is updated using the shared spatial reward and GRPO
objective, while the MedReasoner backbone and MedSAM2 remain frozen. At each
probe event, the updated policy is evaluated again to determine the next domain
sampling distribution.}
\label{fig:overview}
\end{figure*}

\subsection{Coordinating Sampling and Adaptation}

Figure~\ref{fig:overview} shows the two components of \method{}. Let
$\mathcal S_t$ denote the latest probe statistics. For query group $b$ at
update $t$, the scheduler samples
$g_{t,b}\sim p_s(g\mid\mathcal S_t)$ and then draws
$q_{t,b}\sim\mathcal D_{g_{t,b}}$. The known label $g_{t,b}$ also selects
$\phi_{g_{t,b}}$ for rollout generation and optimization. The sampling
distribution changes with the current policy, while adapter selection follows
the observed domain label. Several adapters may be updated in one optimizer
step, but each query group updates only its matching adapter.
The sampling schedule determines which adapters receive updates, and the
updated policy supplies the next probe statistics. This feedback links data
allocation and adaptation throughout training.

\subsection{Estimating Domain Training Value}

We keep probe queries fixed and regenerate their responses with the current
policy to compare domain-level feedback over training. The statistics below
are proxies for training value, not direct estimates of the improvement from
an additional update.
Let $\mathcal V_g$ be the fixed probe-query subset for domain $g$. Every $H$
optimizer updates, the current policy generates $K_p=2$ responses for each
query without gradient computation and with the training-rollout decoding
settings. We record the domain mean reward
$\bar r_g^{(s)}$, mean within-query reward dispersion $u_g^{(s)}$, and valid
output rate $v_g^{(s)}$ at probe stage $s$:
\begin{align}
\bar r_g^{(s)} &=
\frac{1}{|\mathcal V_g|K_p}
\sum_{q\in\mathcal V_g}\sum_{k=1}^{K_p}r_{q,k},\\
u_g^{(s)} &=
\frac{1}{|\mathcal V_g|}
\sum_{q\in\mathcal V_g}\operatorname{Std}_{k}(r_{q,k}),\\
v_g^{(s)} &=
\frac{1}{|\mathcal V_g|K_p}
\sum_{q\in\mathcal V_g}\sum_{k=1}^{K_p}
\mathbb I[o_{q,k}\text{ is valid}].
\end{align}
Here, $\operatorname{Std}_{k}$ denotes the sample standard deviation across
the $K_p$ responses.
An output is valid when it contains the required reasoning and answer blocks
and finite spatial coordinates in the prescribed order.

We first smooth each raw statistic with
\begin{equation}
  \widehat z_g^{(s)}=\eta\widehat z_g^{(s-1)}+(1-\eta)z_g^{(s)},
  \quad z\in\{\bar r,u,v\},\qquad \eta=0.60.
\label{eq:ema}
\end{equation}
We then min--max normalize the smoothed statistics across the $G$ domains.
The raw stage score is
\begin{equation}
  \ell_g^{(s)}=
  \lambda_r\widetilde r_g^{(s)}+
  \lambda_u\widetilde u_g^{(s)}+
  \lambda_v\widetilde v_g^{(s)},
  \qquad
  \lambda_r+\lambda_u+\lambda_v=1.
\label{eq:priority}
\end{equation}
Each tilde denotes the normalized version of the corresponding smoothed
statistic. Finally, the score itself is smoothed before ranking:
\begin{equation}
  \widehat\ell_g^{(s)}=\eta\widehat\ell_g^{(s-1)}+
  (1-\eta)\ell_g^{(s)}.
\label{eq:scoreema}
\end{equation}
Domains are ranked by $\widehat\ell_g^{(s)}$. The normalized progress measure is
$c_g^{(s)}=\operatorname{clip}(\bar r_g^{(s)}/R_{\max},0,1)$ with
$R_{\max}=4$.

Mean reward indicates competence, validity discounts frequent malformed
responses, and dispersion measures the reward contrast available to GRPO.
Admission therefore favors domains that currently provide usable feedback,
rather than simply the hardest examples. Lower-reward domains remain part of
the curriculum: all domains become active in the final stage, where the
deficit term in Eq.~\eqref{eq:allsampling} increases their sampling weight.

\subsection{Scheduling Domain Coverage}

PACE converts the probe scores into a curriculum with three functions:
admitting domains that currently provide usable feedback, revisiting earlier
domains, and ending with coverage of all seven domains.

\subsubsection{Progressive Domain Admission}

PACE maintains an active domain set $\mathcal A_s$ and an inactive set
$\mathcal L_s$. At initialization, the $m_0$ domains with the highest priority
scores form $\mathcal A_0$. The remaining domains are ranked again at each
probe event rather than assigned a fixed order.

We summarize progress on the active set by
\begin{equation}
  C_s=\frac{1}{|\mathcal A_s|}
  \sum_{g\in\mathcal A_s}\widehat c_g^{(s)}.
\end{equation}
Here, $\widehat c_g^{(s)}$ denotes the smoothed normalized reward used by the
competence gate.
After at least one probe in the current stage, the next $m$ locked domains are
admitted if either competence is sufficient or progress has remained on a
plateau:
\begin{equation}
  C_s\ge\tau_c\quad\lor\quad
  \bigcap_{j=0}^{P-1}
  \left\{|C_{s-j}-C_{s-j-1}|<\tau_p\right\}.
\label{eq:gate}
\end{equation}
The first condition waits until the policy reaches a minimum level on the
active domains. The second condition prevents training from stalling when
progress reaches a plateau. When either condition is met, PACE admits the
highest-ranked domains in $\mathcal L_s$ according to the latest probe.

\subsubsection{Replay of Earlier Domains}

Let $\mathcal N_s$ contain the domains newly admitted at stage $s$. Let
$\mathcal R_s=\mathcal A_s\setminus\mathcal N_s$ contain the earlier active
domains. PACE assigns probability mass $1-\rho$ to $\mathcal N_s$ and replay
mass $\rho$ to $\mathcal R_s$. The probability is uniform within each set.
Replay reduces the abrupt sampling shift caused by domain admission. It does
not add examples or change the loss. If no earlier domain exists, the initial
active set receives all probability mass.

\subsubsection{Final All-domain Balancing}

The final checkpoint is evaluated on all domains. PACE therefore reserves the
last fraction of updates for a stage in which every domain is active. The
sampling distribution is
\begin{equation}
p_{\mathrm{all}}^{(s)}(g)=
(1-\alpha)\frac{1}{G}+
\alpha\frac{\exp(\delta_g^{(s)}/\tau)}
{\sum_{j=1}^{G}\exp(\delta_j^{(s)}/\tau)},
\label{eq:allsampling}
\end{equation}
with normalized deficit
\begin{equation}
\delta_g^{(s)}=
\frac{\max_j\widehat r_j^{(s)}-\widehat r_g^{(s)}}
{\max_j\widehat r_j^{(s)}-\min_j\widehat r_j^{(s)}+\epsilon}.
\end{equation}
The uniform component gives every domain non-zero probability. The deficit
component assigns more probability to domains with lower probe reward. This
stage changes only the sampling probability; it does not rescale advantages.

\subsection{Domain-Specific Low-Rank Adaptation}

We replace the single shared LoRA adapter with a bank of lightweight
domain-specific adapters.
For FFN projection $l$ with frozen weight
$W_l\in\mathbb R^{d_o\times d_i}$, domain $g$ owns factors
$A_{l,g}\in\mathbb R^{r\times d_i}$ and
$B_{l,g}\in\mathbb R^{d_o\times r}$. Its forward map is
\begin{equation}
 h_l'=W_lh_l+\frac{\gamma}{r}B_{l,g}A_{l,g}h_l,
 \qquad g=\operatorname{label}(x).
\label{eq:domainmoe}
\end{equation}
We attach the adapters to \texttt{gate\_proj}, \texttt{up\_proj}, and
\texttt{down\_proj} in every language-transformer FFN while excluding the
visual tower. There are seven adapters, one per imaging domain, with rank
$r=16$ and scaling $\gamma=16$. Standard LoRA initialization makes the
initial residual zero, preserving the released model function before RL
updates.

The known modality label stored with each example selects exactly one adapter.
No learned gate or load-balancing loss is required.
At layer $l$, all adapters require $Gr(d_i+d_o)$ stored parameters, whereas a
sample evaluates only $r(d_i+d_o)$ low-rank operations. This is substantially
smaller than copying $G$ dense 7B reasoners and has the active compute order of
one ordinary LoRA adapter.

\subsection{Domain-Conditioned GRPO Updates}

At update $t$, the scheduler independently samples a domain label
$g_{t,b}$ for each query group $b\in\{1,\ldots,B\}$. The query is drawn from
$\mathcal D_{g_{t,b}}$, and the same label activates
$\phi_{g_{t,b}}$ during rollout generation, old-log-probability computation,
and the actor update. Consequently,
\begin{equation}
  \nabla_{\phi_g}\mathcal J_t
  =\sum_{b:g_{t,b}=g}\nabla_{\phi_g}\mathcal J_{t,b};
\label{eq:routedgrad}
\end{equation}
several adapters may receive gradients in one optimizer step. However, each
query group updates only the adapter selected by its domain label. Replay and
the final all-domain stage also revisit adapters that were updated earlier in
training.

\begin{table}[!t]
\centering
\caption{Comparison of domain sampling and parameter-sharing strategies.}
\label{tab:position}
\small
\resizebox{\columnwidth}{!}{%
\begin{tabular}{lccccc}
\toprule
\rowcolor{tablebluehead}
Method & \shortstack{Adaptive\\sampling} & \shortstack{Domain-specific\\adapter} &
\shortstack{Reward\\changed} & Replay & \shortstack{Advantage\\reweighting}\\
\midrule
Vanilla GRPO & \xmark & \xmark & \xmark & \xmark & \xmark\\
Balanced GRPO & \xmark & \xmark & \xmark & \xmark & \xmark\\
Fixed curriculum & \xmark & \xmark & \xmark & \cmark & \xmark\\
DRPO-style & \xmark & \xmark & \xmark & \xmark & \cmark\\
Domain-MoE & \xmark & \cmark & \xmark & \xmark & \xmark\\
PACE-GRPO (Ours) & \cmark & \xmark & \xmark & \cmark & \xmark\\
\rowcolor{tablegreen}
\textbf{\method} & \cmark & \cmark & \xmark & \cmark & \xmark\\
\bottomrule
\end{tabular}}
\end{table}

\section{Experimental Setup}

\subsection{Evaluation Protocol}

We evaluate whether policy-guided domain allocation improves grounding with
domain-specific adapters and whether the sampling benefit also holds with a
shared adapter. The domain-specific comparison uses five seeds (1, 18, 42,
123, and 456); the shared-adapter comparison uses eight seeds. Each comparison
holds the adapter architecture fixed. We then examine alternative sampling
strategies, domain-wise performance, and replay and final-balancing ablations
over seeds 42, 1, and 18. A separate 200-update study tests the effect of a
larger training budget. Cross-architecture differences provide system-level
context because the seed sets, adapter placement, and trainable capacity
differ.

All adapted models start from the released MedReasoner-Soft-7B checkpoint.
The main setting uses 25 optimizer updates; Section~\ref{sec:long} examines
200 updates. We match optimizer updates and gradient-bearing rollouts, not
total generation or wall-clock cost. A 25-update run contains 200
gradient-bearing responses. PACE additionally probes at initialization and
after updates 5, 10, 15, and 20. Each event evaluates 56 held-out queries with
two responses per query, totaling 560 no-gradient responses. A representative
PACE run takes about 65--66 minutes on four GPUs; this timing alone does not
establish a cost advantage over Vanilla.

\subsection{Dataset and Splits}

We use the released category-shift protocol derived from U-MRG
\cite{yan2026medreasoner}. We denote its 6,590-example gradient-bearing update
split as \dataset{}; the probe and test sets are separate held-out partitions
and are not included in the ``6K'' name. The update split contains 2,717 CT,
252 dermoscopy, 134 digital mammography, 417 endoscopy, 2,023 MRI, 254
ultrasound, and 793 X-ray cases. The fixed probe buffer contains 56 examples,
eight per modality. Probe examples guide domain sampling but do not contribute
gradient-bearing responses. The held-out test split contains 1,297 unique
samples and is used for evaluation, not domain scoring.

\subsection{Implementation Details}

All runs start from the downloaded released MedReasoner-Soft-7B checkpoint;
the reference policy and MedSAM2 remain frozen. Each 25-update run uses four
queries per update and two responses per query. Maximum prompt/response lengths
are 2,048/256 tokens, learning rate is $10^{-5}$, and rollouts use temperature
1.0 and top-$p$ 1.0 with top-$k$ disabled. Shared-adapter Vanilla/PACE runs use
the original rank-64 attention LoRA on query, key, value, and output
projections. Domain-MoE runs use seven rank-16 domain-specific FFN LoRA
adapters on gate, up, and down projections. Training uses four GPUs on one
node. All optimizer and GRPO settings not listed below are kept identical to
the released MedReasoner configuration across compared methods.

PACE probes at five-update intervals. It starts with two domains, admits at
most two domains at an expansion event, uses replay ratio $\rho=0.20$, and
reserves the final 20\% of updates for all-domain sampling. The deficit
mixture weight and temperature in Eq.~\eqref{eq:allsampling} are 0.50 and 0.25.
The weights on mean reward, reward dispersion, and validity in
Eq.~\eqref{eq:priority} are 0.45, 0.30, and 0.25, and the EMA decay in
Eq.~\eqref{eq:ema} is 0.60.
% \textbf{Hyper-parameter Details.} For the competence gate
(Eq.~\ref{eq:gate}), we set $\tau_c=0.60$ and $\tau_p=0.01$. A plateau is
declared when $|C_s-C_{s-1}|<\tau_p$ for $P=2$ consecutive probe events.
These values were fixed before the multi-seed comparisons and kept unchanged
across methods and seeds. A systematic sensitivity analysis is left for future
work.
Table~\ref{tab:hyper} summarizes the experimental configuration.

\begin{table}[!t]
\centering
\caption{Configuration of the 25-update experiments.}
\label{tab:hyper}
\small
\resizebox{\columnwidth}{!}{%
\begin{tabular}{ll}
\toprule
Item & Value\\
\midrule
Initialization & Released MedReasoner-Soft-7B\\
Reasoner backbone / segmenter & Lingshu-7B / frozen MedSAM2\\
RL implementation & veRL + vLLM\\
Updates / queries per update & 25 / 4\\
Responses per query & 2\\
Probe responses per query $K_p$ & 2\\
Maximum prompt / response length & 2048 / 256\\
Rollout temperature / top-$p$ / top-$k$ & 1.0 / 1.0 / disabled\\
Learning rate & $1\times10^{-5}$\\
Shared LoRA rank / targets & 64 / attention $q,k,v,o$\\
Domain-specific LoRA rank / targets & 16 / FFN gate, up, down\\
Number of adapters / active adapters & 7 / Top-1\\
Probe interval & 5 updates\\
Probe events / no-gradient responses & 5 / 560\\
Initial / maximum newly admitted domains & 2 / 2\\
Replay ratio $\rho$ & 0.20\\
All-domain fraction / deficit weight $\alpha$ & 0.20 / 0.50\\
Deficit temperature / EMA decay & 0.25 / 0.60\\
Priority-score weights $(\lambda_r,\lambda_u,\lambda_v)$ & $(0.45,0.30,0.25)$\\
Reward maximum $R_{\max}$ & 4.0\\
Competence threshold $\tau_c$ / plateau threshold $\tau_p$ & 0.60 / 0.01\\
Plateau consecutive events $P$ & 2\\
Hardware & 4 GPUs, one node\\
\bottomrule
\end{tabular}
}
\end{table}

\subsection{Baselines and Ablations}

\paragraph{Released Soft-7B}
The public checkpoint is evaluated without additional updates and serves as a zero-update reference.

\paragraph{Vanilla-25}
The released checkpoint is continued for 25 GRPO updates using the original fixed training mixture. This is the primary control.

\paragraph{Balanced-25}
The seven domains are sampled with equal probability throughout training. This tests whether a static correction of domain frequency explains the result.

\paragraph{Fixed-25}
Domains follow the initial probe order: dermoscopy, endoscopy, ultrasound,
digital mammography, X-ray, MRI, and CT. The ranking is frozen, while admission,
replay, and final all-domain balancing remain identical to PACE-25. This
isolates adaptive re-ranking from a fixed ordering.

\paragraph{DRPO-style-25}
Domain-aware scaling is applied after group-relative normalization. The query batch is smaller than in QoQ-Med; we therefore describe this run as DRPO-style rather than as an exact reproduction of its large-batch clustering regime.

\paragraph{PACE-NoReplay-25}
The policy-adaptive domain order is retained, but earlier active domains are not sampled while the curriculum focuses on newly admitted domains.

\paragraph{PACE-NoFinalBalance-25}
Adaptive admission and replay are retained, and all remaining domains are
activated at the reserved final stage. The deficit-weighted distribution in
Eq.~\eqref{eq:allsampling} is disabled. This run isolates the final balancing
strategy while preserving all-domain coverage.

\paragraph{PACE-25}
The full sampler uses adaptive admission, replay, and final all-domain
sampling with one shared LoRA adapter.

\paragraph{MoE-Vanilla-25}
The fixed training mixture is retained, but every sample activates its
rank-16 domain-specific LoRA adapter. This measures the domain-specific
adapter system without adaptive sampling.

\paragraph{PACE-MoE-25}
This is the complete method. Probe statistics determine the domain sampling
distribution, and each known domain label selects its matching LoRA adapter.

\paragraph{Vanilla-200, PACE-200, and PACE-MoE-200}
These seed-42 runs extend post-training to 200 updates from the same released
initialization and are used to compare final grounding performance under the
larger update budget.

\begin{figure*}[!t]
\centering
\includegraphics[width=\textwidth]{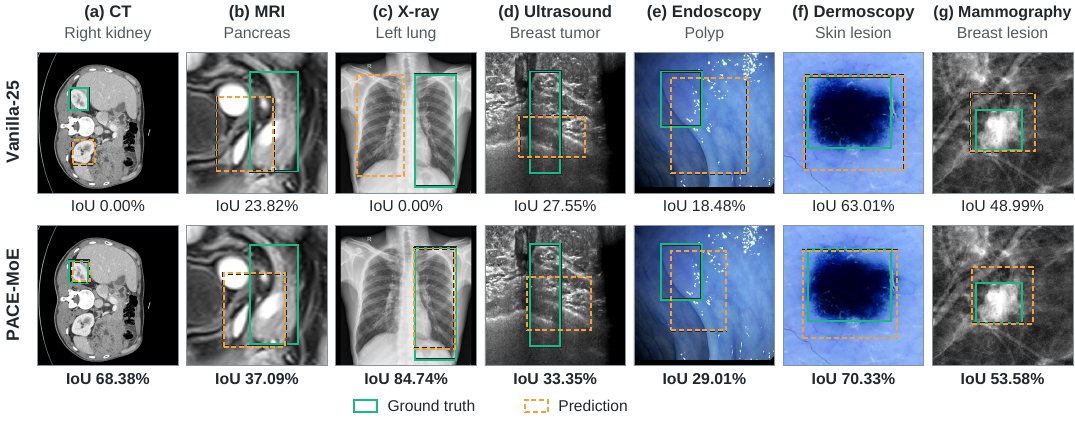}
\caption{Qualitative comparison across seven imaging domains using the same
test image and query for each pair. Top: Vanilla-25. Bottom: \method{}.
Solid green and dashed orange boxes denote ground truth and predictions,
respectively. CT (a) and X-ray (c) retain the full image; the other columns
use identical crops for both methods that retain all compared boxes. IoU is
reported as a percentage and computed in the original image coordinates.}
\label{fig:qualitative}
\end{figure*}

\subsection{Evaluation Metrics}

Box IoU is the primary localization metric. We also report two point-overlap scores. The released MedReasoner evaluator implements
\begin{equation}
  \operatorname{pDice}_{\mathrm{rel}}
  =\frac{2|\mathcal O_p\cap\mathcal O_g|}
  {|\mathcal O_p\cup\mathcal O_g|},
\label{eq:released_pdice}
\end{equation}
where $\mathcal O_p$ and $\mathcal O_g$ are the predicted and reference point
regions. We retain this released-evaluator score for direct comparability.
We also report the conventional Dice form for the same point regions:
\begin{equation}
  \operatorname{pDice}_{\mathrm{std}}
  =\frac{2|\mathcal O_p\cap\mathcal O_g|}
  {|\mathcal O_p|+|\mathcal O_g|}.
\end{equation}
The two scores use different denominators and are reported separately.
Grounding scores are percentages. Relative gains are
$100(S_{\rm method}-S_{\rm base})/S_{\rm base}$, using the named baseline's
reported mean $S_{\rm base}>0$. Paired changes and standard deviations use
one common study-level Vanilla mean. For zero-baseline qualitative cases,
only IoU scores are reported.

The evaluator deduplicates retries by sample identifier. It also verifies the
required reasoning and answer blocks, JSON keys, bounding-box ordering, finite
coordinates, and distinct points before computing spatial metrics. Evaluation
uses deterministic decoding (temperature 0, top-$p$ 1.0, maximum 256 generated
tokens). Reported test results cover all 1,297 unique identifiers. For the
primary eight-seed study, we report the sample standard deviation and two-sided
paired t-test p-values.

\section{Results and Analysis}
\label{sec:results}

\subsection{Grounding Accuracy with PACE-MoE}

Table~\ref{tab:moe_main} compares the complete framework with its controls
after 25 post-training updates. \method{} achieves the highest mean score
on all three grounding metrics. Its mean box IoU is $(36.1182\pm0.1341)\%$,
compared with $(34.8485\pm0.1502)\%$ for Vanilla-25, a relative improvement
of 3.64\%. This comparison evaluates the complete design:
adaptive domain sampling with domain-specific LoRA versus fixed sampling
with a shared LoRA adapter.

\begin{table*}[!t]
\centering
\caption{Multi-seed system comparison and sampling controls after 25 updates. Shared-adapter rows report
eight seeds; domain-specific rows report five seeds (1, 18, 42, 123, and 456).
Grounding scores are mean $\pm$ sample standard deviation, in \%. Cross-architecture rows
compare complete systems; within-architecture pairs assess adaptive sampling.}
\label{tab:moe_main}
\small
\setlength{\tabcolsep}{7pt}
\begin{tabularx}{\textwidth}{>{\raggedright\arraybackslash}Xcrrr}
\toprule
\rowcolor{tablebluehead}
Method & $N$ & Box IoU$\uparrow$ & pDice$_{\rm rel}\uparrow$ & pDice$_{\rm std}\uparrow$\\
\midrule
Released Soft-7B~\cite{yan2026medreasoner} & -- & 34.6734 & 30.4282 & 21.8461\\
Vanilla-25~\cite{yan2026medreasoner} & 8 & $34.8485\pm0.1502$ & $30.5631\pm0.0632$ & $21.8746\pm0.0565$\\
PACE-25 & 8 & $35.2297\pm0.3645$ & $30.8478\pm0.2076$ & $22.1075\pm0.1303$\\
MoE-Vanilla-25 & 5 & $35.3871\pm0.1226$ & $30.9523\pm0.2502$ & $21.9571\pm0.0899$\\
\rowcolor{tablegreen}
\textbf{PACE-MoE-25 (Ours)} & 5 & $\mathbf{36.1182\pm0.1341}$ & $\mathbf{31.6718\pm0.2877}$ & $\mathbf{22.2067\pm0.0929}$\\
\bottomrule
\end{tabularx}
\end{table*}

To assess the sampling contribution separately, we compare \method{} with
MoE-Vanilla-25 using the same seven rank-16 FFN adapters and five seeds.
Adaptive sampling raises mean box IoU from $(35.3871\pm0.1226)\%$ to
$(36.1182\pm0.1341)\%$, a relative gain of 2.07\%. The corresponding relative
gains in pDice$_{\rm rel}$ and pDice$_{\rm std}$ are 2.32\% and 1.14\%.
Thus, adaptive sampling improves grounding beyond using domain-specific
adapters with a fixed data mixture.

These comparisons answer different questions. The 3.64\% system-level gain
includes adapter capacity and placement changes and compares eight-seed
with five-seed means; it does not isolate either component. The 2.07\%
within-architecture gain holds adapters and seeds fixed to assess sampling.
The shared-adapter study below tests the same sampling strategy with a
single adapter.

\subsection{Qualitative Comparison Across Domains}
\label{sec:qualitative}

Figure~\ref{fig:qualitative} illustrates the spatial changes in the
predictions across all seven domains. The clearest differences occur in the
CT and chest X-ray examples. In the CT case (a), the Vanilla-25 prediction
falls on a different kidney region and has no overlap with the annotated
right kidney. \method{} localizes the target with an IoU of 68.38\%. In the
X-ray case (c), Vanilla-25 predicts the opposite lung, whereas \method{}
overlaps the annotated left lung with an IoU of 84.74\%. In both examples,
the change concerns identifying the correct anatomical target, rather than
only adjusting the boundary of an already overlapping box.
The remaining five columns show increased overlap in MRI, ultrasound,
endoscopy, dermoscopy, and mammography. Together, these comparisons provide
visual context for the full-test-set evaluation in
Tables~\ref{tab:moe_main} and~\ref{tab:primary}.

\subsection{Consistency of Adaptive Sampling}

\begin{table}[!htb]
\centering
\caption{Shared-adapter results after 25 updates on 1,297 test cases. Scores (\%): mean $\pm$ sample standard deviation over 8 seeds (1, 18, 36, 42, 123, 456, 999, 2026). Gains are relative to Vanilla-25; paired tests use original scores.}
\label{tab:primary}
\footnotesize
\setlength{\tabcolsep}{1.4pt}
\resizebox{\columnwidth}{!}{%
\begin{tabular}{@{}lrrr@{}}
\toprule
Method & Box IoU$\uparrow$ & pDice$_{\rm rel}\uparrow$ & pDice$_{\rm std}\uparrow$\\
\midrule
Released Soft-7B~\cite{yan2026medreasoner} & 34.6734 & 30.4282 & 21.8461\\
Vanilla-25~\cite{yan2026medreasoner} & $34.8485 \pm 0.1502$ & $30.5631 \pm 0.0632$ & $21.8746 \pm 0.0565$\\
\textbf{PACE-25 (Ours)} & \textbf{$35.2297 \pm 0.3645$} & \textbf{$30.8478 \pm 0.2076$} & \textbf{$22.1075 \pm 0.1303$}\\
\midrule
Relative gain (\%) & $+1.09$ & $+0.93$ & $+1.06$\\
\rowcolor{tableyellowstrong}
Paired $t$-test $p$ & \textbf{0.0023} & \textbf{0.0023} & \textbf{0.0003}\\
\bottomrule
\end{tabular}}
\end{table}

Table~\ref{tab:primary} isolates adaptive domain sampling. Relative to
fixed-mixture training across eight seeds, PACE improves mean box IoU by
1.09\% ($p=0.0023$), $\operatorname{pDice}_{\rm rel}$ by 0.93\%
($p=0.0023$), and $\operatorname{pDice}_{\rm std}$ by 1.06\%
($p=0.0003$). All three remain significant after Holm correction across the
reported metrics (adjusted $p\le0.0046$). Relative to the released checkpoint,
the relative gains in mean IoU are
0.50\% for Vanilla-25 and 1.60\% for PACE-25. These latter differences
are contextual because the released checkpoint is a single, zero-update model
rather than a paired training condition.

\begin{figure*}[t]
\centering
\begin{tikzpicture}
\begin{groupplot}[
  group style={group size=3 by 1, horizontal sep=0.75cm},
  width=0.26\textwidth,
  height=4.4cm,
  xmin=-0.12,
  xmax=1.12,
  xtick={0,1},
  xticklabels={Vanilla-25, PACE-25},
  tick label style={font=\scriptsize},
  x tick label style={rotate=25,anchor=north east},
  title style={font=\small\bfseries},
  ymajorgrids=true,
  grid style={black!12},
  axis line style={black!55},
  every axis plot/.append style={thick, mark=*, mark size=1.8pt},
]
\nextgroupplot[
  title={Box IoU (\%)}, ymin=34.5, ymax=36.0,
  legend style={at={(-0.62,0.50)}, anchor=east, draw=none,
    font=\scriptsize, row sep=-1pt, cells={anchor=west}}
]
% Seed 1
\addplot[seedone] coordinates {(0,34.8109) (1,35.2311)};
\addlegendentry{Seed 1}
% Seed 18
\addplot[seedeighteen] coordinates {(0,34.9723) (1,35.3992)};
\addlegendentry{Seed 18}
% Seed 36
\addplot[seedtsix] coordinates {(0,34.7396) (1,35.1003)};
\addlegendentry{Seed 36}
% Seed 42
\addplot[seedftwo] coordinates {(0,35.0129) (1,35.8298)};
\addlegendentry{Seed 42}
% Seed 123
\addplot[seedotthree] coordinates {(0,34.7124) (1,34.8286)};
\addlegendentry{Seed 123}
% Seed 456
\addplot[seedfsix] coordinates {(0,35.0801) (1,35.6285)};
\addlegendentry{Seed 456}
% Seed 999
\addplot[seednnine] coordinates {(0,34.6990) (1,34.9024)};
\addlegendentry{Seed 999}
% Seed 2026
\addplot[seedttsix] coordinates {(0,34.7608) (1,34.9180)};
\addlegendentry{Seed 2026}

\nextgroupplot[title={pDice$_{\rm rel}$ (\%)}, ymin=30.4, ymax=31.2]
\addplot[seedone] coordinates {(0,30.6028) (1,30.8937)};
\addplot[seedeighteen] coordinates {(0,30.4924) (1,30.9372)};
\addplot[seedtsix] coordinates {(0,30.5739) (1,30.9102)};
\addplot[seedftwo] coordinates {(0,30.5933) (1,31.0832)};
\addplot[seedotthree] coordinates {(0,30.4761) (1,30.6102)};
\addplot[seedfsix] coordinates {(0,30.6629) (1,31.0910)};
\addplot[seednnine] coordinates {(0,30.5873) (1,30.7291)};
\addplot[seedttsix] coordinates {(0,30.5158) (1,30.5277)};

\nextgroupplot[title={pDice$_{\rm std}$ (\%)}, ymin=21.7, ymax=22.5]
\addplot[seedone] coordinates {(0,21.8227) (1,22.1765)};
\addplot[seedeighteen] coordinates {(0,21.8724) (1,22.0883)};
\addplot[seedtsix] coordinates {(0,21.9037) (1,22.1266)};
\addplot[seedftwo] coordinates {(0,21.9745) (1,22.3856)};
\addplot[seedotthree] coordinates {(0,21.8918) (1,22.0799)};
\addplot[seedfsix] coordinates {(0,21.7823) (1,21.9901)};
\addplot[seednnine] coordinates {(0,21.8702) (1,22.0305)};
\addplot[seedttsix] coordinates {(0,21.8789) (1,21.9827)};
\end{groupplot}
\end{tikzpicture}
\caption{Paired seed-level changes from fixed to adaptive domain sampling.
Scores are in \%. Each line connects runs with the same seed and initialization. All 24
metric--seed pairs improve with adaptive sampling.}
\label{fig:paired_seeds}
\end{figure*}
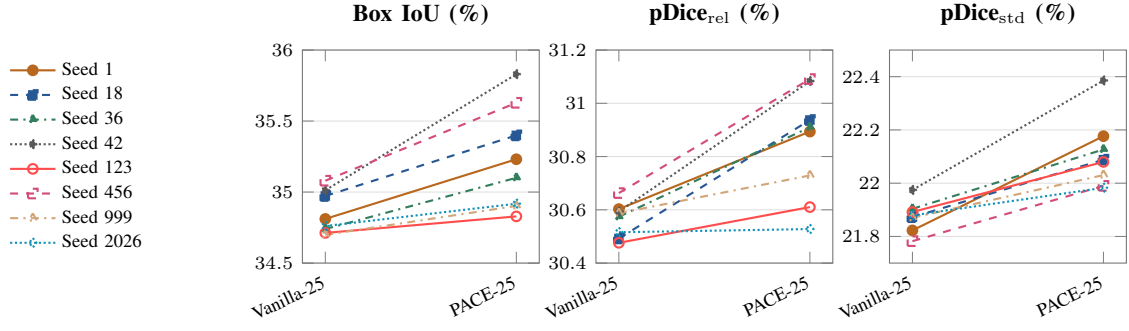

PACE improves box IoU in all eight paired runs
(Fig.~\ref{fig:paired_seeds}). Normalizing each paired change by the common
Vanilla mean of 34.8485\% gives gains from 0.33\% to 2.34\%, with a mean
of 1.09\% and a sample standard deviation of 0.66\%.

\subsection{Why Adapt the Domain Schedule?}

\begin{table}[!t]
\centering
\caption{Seed-42 control study. Grounding scores are in \%. Every adapted method starts from the same checkpoint and uses 25 optimizer updates.}
\label{tab:controls}
\scriptsize
\resizebox{\columnwidth}{!}{%
\begin{tabular}{lrrr}
\toprule
\rowcolor{tablebluehead}
Method & Box IoU$\uparrow$ & pDice$_{\rm rel}\uparrow$ & pDice$_{\rm std}\uparrow$\\
\midrule
\rowcolor{tablegray}
Released Soft-7B~\cite{yan2026medreasoner} & 34.6734 & 30.4282 & 21.8461\\
\rowcolor{tableblue}
Vanilla-25~\cite{yan2026medreasoner} & 35.0129 & 30.5933 & 21.9745\\
\rowcolor{tableblue}
Balanced-25 & 34.9135 & 30.6818 & 22.0257\\
\rowcolor{tableblue}
Fixed-25 & 34.7962 & 30.7027 & 22.0253\\
\rowcolor{tableblue}
DRPO-style-25~\cite{dai2025qoqmed} & 34.6774 & 30.7705 & 21.8747\\
\rowcolor{tableyellow}
PACE-NoReplay-25 & 34.8871 & 30.8690 & 21.9824\\
\rowcolor{tableyellow}
PACE-NoFinalBalance-25 & 34.7647 & 30.5863 & 21.9337\\
\rowcolor{tableyellowstrong}
\textbf{PACE-25  (Ours)} & \textbf{35.8298} & \textbf{31.0832} & \textbf{22.3856}\\
\bottomrule
\end{tabular}}
\end{table}

\begin{figure}[!tb]
\centering
\begin{tikzpicture}
\begin{axis}[
  width=0.92\columnwidth,
  height=4.6cm,
  xbar,
  bar width=7pt,
  bar shift=0pt,
  xmin=-0.70,
  xtick={0,2,4,6,8,10,12},
  xmax=12.3,
  xlabel={Relative Box-IoU gain (\%)},
  symbolic y coords={MRI,Dermoscopy,X-ray,Endoscopy,CT,Ultrasound,Mammography},
  ytick={MRI,Dermoscopy,X-ray,Endoscopy,CT,Ultrasound,Mammography},
  tick label style={font=\scriptsize},
  label style={font=\scriptsize},
  xmajorgrids=true,
  grid style={black!12},
  axis line style={black!55},
  nodes near coords={\pgfmathprintnumber[fixed,precision=2]{\pgfplotspointmeta}},
  point meta=x,
  every node near coord/.append style={font=\scriptsize,anchor=west},
]
\addplot[fill=paceblue!72,draw=paceblue]
  coordinates {(4.335951,CT) (1.901017,Dermoscopy) (10.213551,Mammography)
  (2.587954,Endoscopy) (4.628277,Ultrasound) (2.251692,X-ray)};
\addplot[fill=crossred!65,draw=crossred]
  coordinates {(-0.033258,MRI)};
\end{axis}
\end{tikzpicture}
\caption{Seed-42 relative box-IoU gains by imaging domain, using each
domain's Vanilla-25 score as the baseline. Six domains improve; MRI changes
by $-0.03\%$.}
\label{fig:domain_delta}
\end{figure}
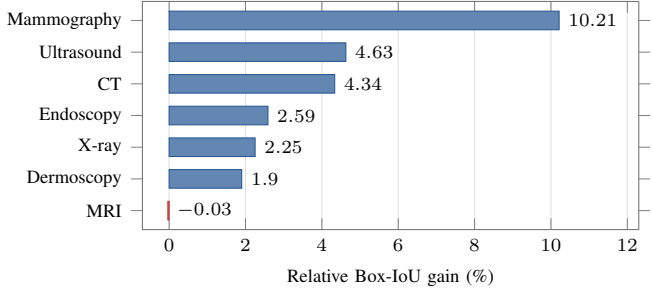

The seed-42 controls answer three narrower questions. First, equal domain
sampling does not reproduce the PACE result: PACE improves box IoU by 2.62\%
relative to Balanced-25. Second, a fixed domain order is also insufficient;
PACE improves by 2.97\% relative to Fixed-25. Third, advantage reweighting is
not equivalent to changing which domains are sampled: PACE improves by
3.32\% relative to DRPO-style-25.

Taken together, the matched seed-42 controls show that neither static domain balancing, fixed ordering, nor post-sampling advantage scaling reproduces the full PACE checkpoint.

\subsection{Domain-Wise Grounding Performance}
\label{sec:domain}

The aggregate score can conceal whether scheduling benefits a narrow subset of
modalities or shifts error from one domain to another. We therefore decompose
the matched seed-42 predictions by the seven test domains. The same 1,297
sample identifiers and the same evaluator are used for both checkpoints.
\begin{table}[!t]
\centering
\caption{Seed-42 domain-level box IoU (\%) on the 1,297-case test split.
Gain is the relative change from Vanilla-25, in \%. The macro average weights the
seven domains equally, whereas the overall result is sample-weighted.}
\label{tab:domain_breakdown}
\small
\resizebox{\columnwidth}{!}{%
\begin{tabular}{lrrrr}
\toprule
\rowcolor{tablebluehead}
Imaging domain & $N$ & Vanilla-25 & PACE-25 & Gain (\%)\\
\midrule
CT & 528 & \cellcolor{tableblue}25.5192 &
\cellcolor{tableyellow}\textbf{26.6257} & \textbf{+4.34}\\
Dermoscopy & 51 & \cellcolor{tableblue}55.7754 &
\cellcolor{tableyellow}\textbf{56.8357} & \textbf{+1.90}\\
Mammography & 23 & \cellcolor{tableblue}22.3928 &
\cellcolor{tableyellow}\textbf{24.6799} & \textbf{+10.21}\\
Endoscopy & 85 & \cellcolor{tableblue}44.5062 &
\cellcolor{tableyellow}\textbf{45.6580} & \textbf{+2.59}\\
MRI & 397 & \cellcolor{tableblue}\textbf{36.6831} &
  \cellcolor{tableyellow}36.6709 & \textcolor{crossred}{-0.03}\\
Ultrasound & 53 & \cellcolor{tableblue}36.1560 &
\cellcolor{tableyellow}\textbf{37.8294} & \textbf{+4.63}\\
X-ray & 160 & \cellcolor{tableblue}51.9698 &
\cellcolor{tableyellow}\textbf{53.1400} & \textbf{+2.25}\\
\midrule
Macro average & -- & \cellcolor{tableblue}39.0004 &
\cellcolor{tableyellow}\textbf{40.2056} & \textbf{+3.09}\\
Overall & 1,297 & \cellcolor{tableblue}35.0129 &
\cellcolor{tableyellowstrong}\textbf{35.8298} & \textbf{+2.33}\\
\bottomrule
\end{tabular}}
\end{table}
PACE improves box IoU in six of the seven domains
(Table~\ref{tab:domain_breakdown} and Fig.~\ref{fig:domain_delta}). The largest
relative gains occur on mammography (10.21\%) and ultrasound (4.63\%),
followed by CT (4.34\%), endoscopy (2.59\%), X-ray (2.25\%), and dermoscopy
(1.90\%). MRI is the only exception, changing from 36.6831\% to 36.6709\%,
a relative decrease of 0.03\%. The macro-average score improves by 3.09\%
relative to the Vanilla macro average, compared with a 2.33\% improvement
in the sample-weighted score. Some of the largest gains occur in smaller
domains. At the same time, CT contains 528 test cases and improves by 4.34\%,
so the aggregate result is not produced
solely by rare modalities. These values are descriptive results from the
matched seed-42 pair and are not used for domain-level significance claims.

\subsection{Contribution of Replay and Final Balancing}

\begin{table}[!htb]
\centering
\caption{PACE ablations after 25 updates across seeds 42, 1, and 18. Grounding scores are mean $\pm$ standard deviation, in \%.}
\label{tab:ablation_summary}
\footnotesize
\setlength{\tabcolsep}{1.8pt}
\renewcommand{\arraystretch}{1.30}
\begin{tabularx}{\columnwidth}{@{}l*{3}{>{\raggedleft\arraybackslash}X}@{}}
\toprule
\rowcolor{tablebluehead}
Method & Box IoU$\uparrow$ & pDice$_{\rm rel}\uparrow$ & pDice$_{\rm std}\uparrow$ \\
\midrule
Vanilla & 34.9320\,${\pm}$\,0.1069 & 30.5628\,${\pm}$\,0.0612 & 21.8899\,${\pm}$\,0.0774 \\
w/o replay & 34.9364\,${\pm}$\,0.0970 & 30.7735\,${\pm}$\,0.0837 & 21.8937\,${\pm}$\,0.0775 \\
w/o final balance & 34.9302\,${\pm}$\,0.1692 & 30.6962\,${\pm}$\,0.1120 & 21.8361\,${\pm}$\,0.0855 \\
\textbf{PACE (full)} & \textbf{35.4867\,${\pm}$\,0.3088} & \textbf{30.9714\,${\pm}$\,0.0993} & \textbf{22.2168\,${\pm}$\,0.1527} \\
\bottomrule
\end{tabularx}
\end{table}

\begin{figure}[!htb]
\centering
\begin{tikzpicture}
\begin{axis}[
  width=\columnwidth,
  height=4.3cm,
  ybar,
  bar width=18pt,
  ymin=-0.90,
  label style={font=\scriptsize},
  ymax=3.10,
  ylabel={Relative Box-IoU gain (\%)},
  symbolic x coords={NoReplay,NoFinal,PACE},
  xtick={NoReplay,NoFinal,PACE},
  x tick label style={rotate=25,anchor=east,font=\scriptsize},
  y tick label style={font=\scriptsize},
  enlarge x limits=0.15,
  ymajorgrids=true,
  grid style={black!12},
  axis line style={black!55},
  error bars/y dir=both,
  error bars/y explicit,
]
\addplot[fill=paceblue!68,draw=paceblue]
  coordinates {(NoReplay,0.012310) +- (0,0.323199)
  (NoFinal,-0.005153) +- (0,0.611474)
  (PACE,1.587942) +- (0,0.650120)};
\draw[black!55,dashed] (axis cs:NoReplay,0) -- (axis cs:PACE,0);
% Place horizontal value labels above the complete error bars so that the
% annotations never collide with the whiskers or the plot frame.
\node[font=\scriptsize,anchor=south] at (axis cs:NoReplay,0.415) {$+0.012\%$};
\node[font=\scriptsize,anchor=south] at (axis cs:NoFinal,0.701) {$-0.005\%$};
\node[font=\scriptsize,anchor=south] at (axis cs:PACE,2.333) {$+1.588\%$};
\end{axis}
\end{tikzpicture}
\caption{Paired ablations across seeds 42, 1, and 18: mean $\pm1$ sample
standard deviation of paired box-IoU changes, both normalized by the
common Vanilla-25 mean (34.9320\%) and expressed as percentages.}
\label{fig:ablation_summary}
\end{figure}
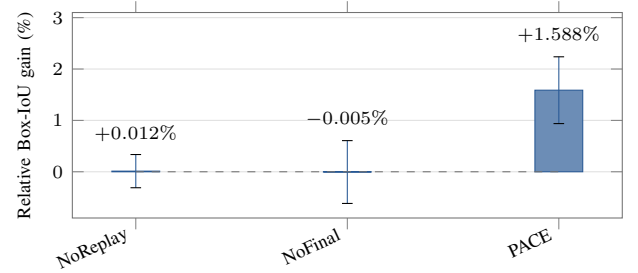

Across three random seeds (42, 1, 18), we repeat the ablation study to assess
the contribution of replay and final balancing. Table~\ref{tab:ablation_summary}
and Fig.~\ref{fig:ablation_summary} summarize the results.
Across seeds 42, 1, and 18, full PACE improves mean box IoU over Vanilla-25
by 1.59\%. Normalizing paired changes by the common Vanilla mean gives a
range of 1.20\%--2.34\%. PACE-NoReplay obtains a mean IoU of 34.9364\%, and
PACE-NoFinalBalance obtains 34.9302\%. Both values are close to the
Vanilla-25 mean of 34.9320\%.
The ablation pattern is not identical in every seed. Both variants slightly
exceed Vanilla-25 in seeds 1 and 18 but fall below it in seed 42. The magnitude
of the component effect therefore depends on the seed, while full PACE remains
above both variants in all three matched comparisons. This consistent
ordering supports retaining replay and final balancing together in the
tested schedule.

\subsection{Effect of Additional Post-Training Updates}
\label{sec:long}

\begin{table}[!htb]
\centering
\caption{Seed-42 extended-training analysis after 200 post-training updates. Grounding scores are in \%.}
\label{tab:long}
\small
\begin{tabular}{lrrr}
\toprule
Method & IoU$\uparrow$ & pDice$_{\rm rel}\uparrow$ & pDice$_{\rm std}\uparrow$\\
\midrule
Released Soft-7B~\cite{yan2026medreasoner} & 34.6734 & 30.4282 & 21.8461\\
Vanilla-200~\cite{yan2026medreasoner} & 34.4138 & 30.2541 & 22.0571\\
PACE-200 & 34.5867 & 30.3731 & 21.8042\\
PACE-MoE-200 & 34.6691 & 30.3822 & 21.9208\\
\bottomrule
\end{tabular}
\end{table}

At 200 updates, PACE-MoE achieves the highest final box IoU among the three
continuation methods (Table~\ref{tab:long}). Its relative gains are 0.24\%
over PACE-200 and 0.74\% over Vanilla-200. Its box IoU of 34.6691\% is close
to the released checkpoint at 34.6734\%, whereas the other continuation methods
finish further below that reference. This endpoint does not improve on the
released model and is below the five-seed mean observed after 25 updates.
The 200-update result uses only seed 42, so the two summaries do not form a
paired training curve. The result shows that the reported gains cannot be
assumed to persist with more updates; it does not establish a convergence
rate or a forgetting mechanism.

\section{Discussion}

\subsection{What Does Policy Feedback Contribute?}

The full-system comparison evaluates the combined design, whereas the
within-architecture comparisons assess adaptive sampling without changing
adapter placement or capacity. The latter show a positive mean sampling
gain in both adapter architectures. The fixed-order control retains admission, replay, and final
balancing, but freezes the initial ranking. Its lower seed-42 result suggests
that updating the order matters beyond merely dividing training into stages.
The ablations also show that prioritization alone is insufficient in the
tested setting. Replay keeps earlier domains in the sampling distribution,
while final balancing increases the probability of lower-reward domains.
Removing either component brings the three-seed mean close to Vanilla-25.
The practical role of the complete schedule is therefore to combine selective
training with continued domain coverage. This observation supports the design
without identifying catastrophic forgetting as the cause of an ablation's
lower score.

\subsection{Domain-Specific Adaptation and Training Cost}

Sampling controls how often each domain is trained; the known label selects
the parameters that receive its updates. Neither decision requires changing
the GRPO advantage or reward. The design shares one frozen backbone and
activates one adapter per sample, although all seven adapters must be stored.
Probe generation also adds computation. Consequently, the experiments support
better grounding at a fixed optimizer-update budget, not a reduction in
total generation or wall-clock cost.

\subsection{Limitations}

The evidence covers one benchmark and one backbone. The shared-adapter and
domain-specific comparisons use eight and five seeds, respectively, but do
not provide a capacity- and placement-matched estimate of the adapter effect.
The domain-wise aggregate analysis is available for the shared-adapter pair,
not for the complete \method{} system. The probe size and scheduler
hyperparameters also lack systematic sensitivity analysis. Hard routing
assumes correct imaging-domain labels; missing, mixed, and noisy labels are
not evaluated. The experiments do not include compute-matched controls or
direct implementations of recent bandit and prioritized-replay samplers.
Finally, box and point grounding do not measure final mask quality or clinical
utility. Broader validation should examine these factors and training budgets
beyond the repeated-seed 25-update setting.

\section{Conclusion}

We presented \method{} for domain-aware medical visual grounding. Its core
idea is to use policy feedback to allocate training queries and the known
domain label to assign their updates to domain-specific adapters. This
coordinates data allocation and model adaptation while preserving shared
pretrained representations and the original GRPO objective. Across seven
imaging domains, the complete method reaches 36.12\% mean box IoU after 25
updates, a 3.64\% relative improvement over Vanilla with fixed sampling and
a shared adapter. With the domain-specific adapter architecture held fixed,
adaptive sampling provides a 2.07\% relative gain over fixed sampling; a positive sampling
gain also holds with shared adapters. Ablations support combining adaptive
admission with replay and final balancing. These findings motivate further
study of coordinated domain sampling and adaptation across backbones,
datasets, and training budgets.

% \section*{Compliance Statements}

% This study uses the public U-MRG benchmark and involves no prospective data
% collection or clinical decision-making. We collected no new clinical data and
% declare no competing interests.

\footnotesize
% Balance the final reference page with the standard IEEEtran hook.
\IEEEtriggeratref{31}

\end{document}